\documentclass[sn-mathphys-num]{sn-jnl}

\usepackage{graphicx}%
\usepackage{multirow}%
\usepackage{amsmath,amssymb,amsfonts}%
\usepackage{amsthm}%
\usepackage{mathrsfs}%
\usepackage[title]{appendix}%
\usepackage{xcolor}%
\usepackage{textcomp}%
\usepackage{manyfoot}%
\usepackage{booktabs}%
\usepackage{algorithm}%
\usepackage{algorithmicx}%
\usepackage{algpseudocode}%
\usepackage{listings}%

\newcommand{\ignore}[1]{}

\theoremstyle{thmstyleone}%
\theoremstyle{thmstyletwo}%

\theoremstyle{thmstylethree}%

\begin{document}

\title[Data Augmentation with In-Context Learning and Comparative Evaluation in Math Word Problem Solving]{Data Augmentation with In-Context Learning and Comparative Evaluation in Math Word Problem Solving}

\author*[1,2]{\fnm{Gulsum} \sur{Yigit}}\email{gulsum.yigit@std.yildiz.edu.tr}
\equalcont{These authors contributed equally to this work.}

\author[1]{\fnm{Mehmet Fatih} \sur{Amasyali}}\email{amasyali@yildiz.edu.tr}
\equalcont{These authors contributed equally to this work.}

\affil*[1]{\orgdiv{Department of Computer Engineering}, \orgname{Yildiz Technical University}, \orgaddress{\city{Istanbul}, \country{Turkey}}}
\affil[2]{\orgdiv{Department of Computer Engineering}, \orgname{Kadir Has University}, \orgaddress{\city{Istanbul},  \country{Turkey}}}


\abstract{

Math Word Problem (MWP) solving presents a challenging task in Natural Language Processing (NLP). This study aims to provide MWP solvers with a more diverse training set, ultimately improving their ability to solve various math problems. We propose several methods for data augmentation by modifying the problem texts and equations, such as synonym replacement, rule-based: question replacement, and rule based: reversing question methodologies over two English MWP datasets. This study extends by introducing a new in-context learning augmentation method, employing the Llama-7b language model. This approach involves instruction-based prompting for rephrasing the math problem texts. 
Performance evaluations are conducted on 9 baseline models, revealing that augmentation methods outperform baseline models. Moreover, concatenating examples generated by various augmentation methods further improves performance.
}




\maketitle

\section{Introduction} \label{sec:introduction}

Question Answering (QA) systems are instrumental in Natural Language Processing (NLP). QA systems are responsible for understanding and reacting to the user questions in a manner resembling a human \cite{yigit2023enhancing, wu2020seq2seq, fan2019eli5, jin2023building}. This makes them essential for various applications such as search engines, virtual assistants, information retrieval systems, etc. 
Enhancing these systems is critical as the demand for instant and accurate information retrieval continues.
The structure of QA systems varies based on the domain and the types of questions they have \cite{abdel2023deep, rogers2023qa, yigit2019ask}. One important form is solving Math Word Problems (MWPs). 

MWPs are a complex category within the field of QA systems
\cite{xie2019goal, zhang2020teacher,liang2022mwp, wang2019template, zhang2020graph, shen2020solving}. Challenges in these systems require knowledge beyond simple pattern recognition and keyword matching. These special systems involve understanding given mathematical operators, quantities, and their complex relationships with each other. MWPs aim to produce a solution equation. Therefore, it requires identifying numerical values provided in context, carefully selecting appropriate mathematical operations, and transforming them into particular mathematical expressions involving unknown variables.

Various datasets have been produced in MWPs with unique features and purposes to benefit research and growth in this field. These datasets vary in source, complexity, scale, and the number of unknowns. For example,  Draw1K \cite{upadhyay2016annotating}, HMWP \cite{qin2020semantically}, ALG514 \cite{kushman2014learning}, MAWPS \cite{koncel2016mawps}  emphasize complex multi-unknown word problems, while SVAMP \cite{patel2021nlp}, MAWPS-Single \cite{koncel2016mawps}, ASDIV-A \cite{miao2021diverse}, Math23K \cite{wang2017deep}  focus on elementary math questions with a single unknown.
Despite their differences, these datasets share common challenges, such as the need for semantic understanding, extracting numerical information, and changing textual context into mathematical expressions. Addressing these challenges is essential for developing practical solutions in MWP scenarios, highlighting the significance of these datasets in advancing the state-of-the-art of MWP systems.

MWPs include a variety of approaches. Statistical methods often use statistical patterns and correlations within the dataset \cite{roy2018mapping, mitra2016learning}. Rule-based approaches involve applying predefined rules to solve MWPs  \cite{fletcher1985understanding, bakman2007robust, yuhui2010frame, kushman2014learning}. Semantic parsing methods attempt to understand the underlying meaning of the text and transform it into mathematical expressions. On the other hand, deep learning-based models use neural networks to extract contextual information and capture complex relationships between textual context and mathematical concepts \cite{wang2017deep, wang2018translating}.
Moreover, there are MWP solvers that have been developed to include pre-trained models. Pre-trained models such as BERT, GPT and their variants have gained great importance recently. These models are first pre-trained on massive amounts of textual data, which used to capture complex linguistic patterns and contextual information. When fine-tuned to solve MWP, they show significant improvements in understanding the language used in MWPs. These different approaches enable a more comprehensive exploration of potential solutions in the field of MWPs.

Data augmentation in NLP is essential to improve the performance and robustness of the models. It helps to increase model robustness and reduce overfitting by providing multiple variations of existing data. However, challenges arise, especially in maintaining semantics. Simple transformations on textual data can change the meaning of the original context. 
There are other significant challenges to implementing data augmentation in an MWP dataset. One issue is model bias, where augmented data tends towards specific problems. This leads to inaccurate predictions and biased results.
Processing numerical values in the generated augmented samples introduces another problem: the augmentation procedure may produce inconsistent numerical information. Additionally, modifying the original data distribution through data augmentation increases the likelihood of overfitting or underfitting for specific problems. Factors such as data sparsity issues and computational overhead contribute to the difficulties in data augmentation in the MWP datasets.

In this study, our primary focus has been on enriching MWP datasets via useful data augmentation methods. We aim to augment training data by modifying the source text and equations. Building on our previous study in \cite{yigit2023exploring}, we introduced three approaches for data augmentation:
Rule-Based: Question Replacement, Rule-Based: Reversing Question and Substitution: Synonym Replacement.
Besides, we extended our work by presenting a novel in-context learning augmentation method, leveraging the Llama-7b language model \cite{touvron2023llama}. This method employs instruction-based prompting for rephrasing problem texts. The model generates a rephrased version for each training example, resulting in new samples after filtering and numeric modification steps. 
In this study, we have the following contributions.
\begin{itemize}
    \item We proposed several augmentation methods: Rule-Based: Question Replacement, Rule-Based: Reversing Question, and Substitution: Synonym Replacement. Besides, we extended our work by presenting a novel in-context learning augmentation method, leveraging the Llama-7b language model.
    \item We also extended prior experiments and tested these augmentation methods on 9 different baseline models using the MWPS-Single and SVAMP datasets. By comparing the results with the baseline of an earlier study \cite{lan2022mwptoolkit}, we found that these augmentation methods consistently lead to improved performance in MWP solving.
\end{itemize}

This paper is structured as follows: Section \ref{sec:related_work} provides a review of MWP solving systems. Section \ref{sec:proposed} delves into the introduced augmentation methods. In Section \ref{sec:experiments}, detailed information about the experiments is presented. We performed comprehensive experiments on 2 MWP datasets and evaluated 9 different models. Section \ref{sec:discussion} provides a discussion about approaches and Section \ref{sec:conclusion} concludes the paper.

\section{Related Work} \label{sec:related_work}
The investigation of automatic MWP solvers has attracted considerable attention by operating various approaches. 
One early method is the rule-based approach, where decisions are based on a predefined set of rules. This method maps equations into templates, extracting predefined patterns from the problem texts \cite{fletcher1985understanding, bakman2007robust, yuhui2010frame, kushman2014learning}. 
Studies using rule-based systems have demonstrated satisfactory performances, specifically on small-scale MWP datasets. Nevertheless, a noteworthy disadvantage of this method is its dependency on human intervention to create templates. 
Preparing effectual templates requires a profound understanding of problem structures. It makes the rule-based systems challenging as the complexity of the problems increases. The work in \cite{yuhui2010frame} presents MSWPAS-CP, a computer simulation system developed to help students solve multi-step arithmetic word problems. The system processes natural language word problems into frames. It performs calculus based on these frames.

Another method in earlier investigations concerns a statistical-based procedure, where a statistical classifier makes decisions. This approach depends on analyzing statistical patterns to make predictions \cite{hosseini2014learning, roy2018mapping, zhou2015learn}. The study in \cite{zhou2015learn} introduces an algorithm for automatically solving algebra word problems. It analyzes a hypothesis space that contains all potential equations derived from assigning numbers. Via training a log-linear model to optimize the margin between correct and false assignments, the algorithm efficiently addresses a quadratic programming problem. 
Moreover, some earlier works apply semantic parsing \cite{koncel2015parsing, zhou2015learn, huang2017learning}. Zhou et al. introduced a semantic parsing and reasoning approach that utilizes a new meaning representation language (DOL) to connect natural language text and mathematical expressions, employing a parser to convert text into DOL trees \cite{zhou2015learn}.

Research has moved towards utilizing deep learning-based approaches to reduce human intervention and improve the MWP solvers. One well-known method concerns using Sequence-to-Sequence (Seq2Seq) models to enhance the performance of MWP solvers \cite{wang2017deep, zhang2016variational, wang2018translating, huang2018neural}. 
Huang et al. suggested incorporating a copy-and-alignment mechanism into the traditional Seq2Seq model \cite{huang2018neural}. 
Notably, in a study by Wang et al., Recurrent Neural Networks (RNNs) were utilized to transform MWPs into solution equations \cite{wang2017deep}. This usage of Seq2Seq models seeks to follow the sequential nature of language and problem-solving structures.
Similarly, in \cite{wang2018translating}, the approach comprises equation normalization to handle the challenge of duplicate equations. However, one disadvantage of Seq2Seq models is their potential to yield invalid solution equations due to a lack of management over the decoder during the generation process. The decoder, accountable for constructing the output sequence, may generate solutions that may not be valid in the context of the given problem. This restriction underlines the need to distill Seq2Seq models to create proper and contextually appropriate solution equations.

Some studies have considered integrating expression trees to address the issue of generating valid solution equations using Seq2Seq models. These models are generalized as Seq2Tree models \cite{xie2019goal, wang2019template}. These models go beyond the conventional Seq2Seq architectures, which construct the output as an expression tree, seeking more specific and contextually relevant solution equations. In a study by Xie et al., the authors utilized a Long Short-Term Memory (LSTM) encoder to encode the problem text and presented a tree-based decoder to generate the equation expression \cite{xie2019goal}. This Seq2Tree model denotes a more hierarchical and structured representation of solution equations.
A template-based model was presented in another approach proposed in \cite{wang2019template}. This model interests constructing a Seq2Seq model to predict the tree-structured template for the solution equation. The predicted numeric values are depicted as leaf nodes, while the operators perform as non-leaf nodes in the solution expression tree. This Seq2Tree model provides a more structured outcome and mitigates the case of generating invalid or contextually improper solution equations. Various other models, such as Chiang et al. with a semantic tracking stack \cite{chiang2018semantically}, Li et al. \cite{li2019modeling} incorporating different functional multihead attentions, and Meng et al. \cite{meng2019solving} applying double sequence-based decoders, have been explored.

MWP solvers have seen an expansion towards Graph2Tree approaches \cite{zhang2020graph, shen2020solving}. The Graph2Tree method parses the problem text to construct binary trees that preserve numerical data and mathematical operators. 
Li et al. combined the dependency parse tree and constituency tree from text descriptions, while Zhang et al. developed the quantity cell and comparison graphs \cite{zhang2020graph, li2020graph}. These approaches include structural information from text descriptions.
Shen et al. applied a hybrid technique, integrating a sequence-based encoder with a graph-based decoder \cite{shen2020solving}. This fusion is proposed to improve text representations and generate various solution equations. Incorporating a graph-based decoder permitted the model to offer a more comprehensive understanding and enabled the generation of varied and contextually appropriate solution equations.

Beyond Graph2Tree strategies, some investigators have examined the integration of pre-trained language models \cite{devlin2018bert, liu2019roberta, liang2022mwp, radford2019language}. Liang et al. presented MWP-BERT, leveraging the capabilities of pre-trained language models tailored explicitly for MWP solvers \cite{liang2022mwp}. This approach emphasizes the contextual understanding grabbed by pre-trained models, achieving higher performances by producing more accurate and contextually relevant solution equations.


The emergence of Large Language Models (LLMs) has prompted a surge in innovative approaches to solving MWPs. 
Many researchers have made notable contributions to this field, as evidenced by significant advancements \cite{shao2022chaining, li2022advance, wang2022self, pi2022reasoning, chen2022program, liang2023let}. 
Prompt-based learning has gained attraction for its ability to leverage LLMs' inherent prediction capabilities. Unlike traditional supervised learning, prompt-based learning utilizes text prompts to guide the model's responses, often in a few-shot or zero-shot format. Lazaridou et al. and Chen et al. have explored the optimization of prompts for different tasks, facilitating the generation of desired responses from large models without requiring extensive fine-tuning \cite{lazaridou2022internet, chen2022program}.

Furthermore, Chain of Thought (CoT) prompting has emerged as a promising approach, generating reasoning steps to guide the model in deriving the actual answer. This method not only allows the model to arrive at reasonable conclusions but also enhances its explainability. Wei et al. demonstrated the effectiveness of CoT prompting compared to conventional prompt methods in MWPs \cite{wei2022chain}. 
Li et al. presented DIVERSE (Diverse Verifier on Reasoning Step) as a novel approach to enhance language models' reasoning capability \cite{li2022advance}. It contains three main parts: diverse prompts generation to investigate different reasoning paths for the same question, utilization of a verifier to filter out incorrect answers based on a weighted voting scheme, and verification of each reasoning step individually instead of the entire chain.
Wang et al. presented a novel decoding strategy, self-consistency by sampling diverse reasoning paths and selecting the most consistent answer. Empirical evaluations across various arithmetic reasoning benchmarks show that self-consistency significantly improves the performance of CoT prompting \cite{wang2022self}.

Chen et al. introduced a ``Program of Thoughts" (PoT), which employs language models, primarily Codex, to generate both text and programming language statements, culminating in an answer \cite{chen2022program}. PoT underwent evaluation in both few-shot and zero-shot settings, revealing improvements in performance.
The authors highlight significant limitations observed in CoT, including the propensity of LLMs to encounter arithmetic errors, particularly with large numbers, the challenge faced by LLMs in solving complex mathematical expressions such as polynomial or differential equations, and the inefficiency of LLMs in expressing iteration, especially with numerous steps. In PoT, computation is delegated to an external language interpreter, while reasoning steps are articulated as Python programs by Language Models.

\section{Proposed Augmentation Methods} \label{sec:proposed}

\subsection{Rephrase with In-Context Learning}

In the previous study, an investigation into the impact of paraphrasing models on MWPs highlighted challenges in generating rephrased texts while preserving the integrity of numerical quantities and their relationships \cite{yigit2023exploring}. In this study, we present a new method that introduces a sophisticated approach employing in-context learning to create rephrased examples for MWP datasets.
This approach follows the following three steps.
\begin{itemize}
    \item \textbf{Step 1: In-Context Learning:}
    In the initial step of the proposed method, 15 examples from an MWP dataset are presented to the Llama-7b language model \cite{touvron2023llama}, a powerful tool for natural language understanding. The 15 MWP samples were randomly selected. No specific heuristic was employed in the selection process. Then, the instruction as given below was followed by 15 examples and the rephrased text of these examples. ChatGPT is used to generate the rephrased texts for these 15 examples \cite{brown2020language}. The model is instructed to generate a rephrased version of each training example. The objective is to leverage in-context learning to produce texts while preserving the inherent mathematical structure. 
        \begin{figure}[h]
        \centering
        \includegraphics[scale=0.45]{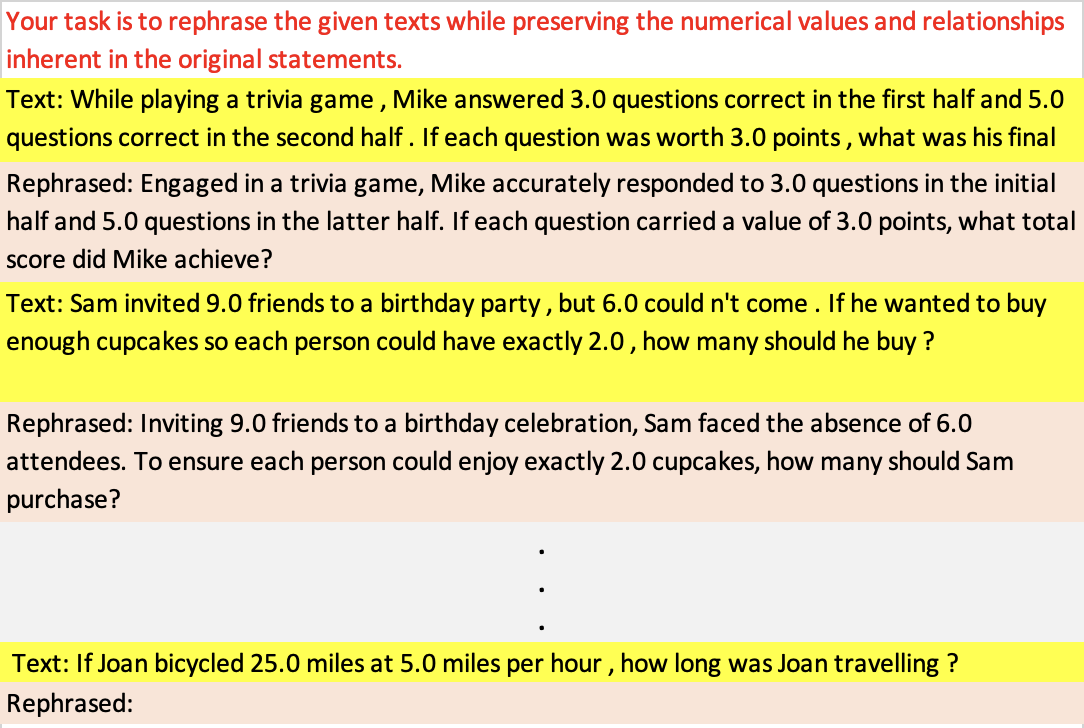}
        \end{figure}
    \item \textbf{Step 2: Filtering Mechanism:} 
    Following the generation process, a specific filtering mechanism is implemented to pick the appropriate ones from the rephrased examples. This process involves two key aspects:
    \begin{itemize}
        \item Validity Check: 
        The filtering mechanism excluded samples lacking a ``?" to ensure they represent valid questions in line with MWPs structures. However, this may not always catch all the valid questions, especially when problem statements end with phrases like ``Find the value of ...". While ``?" serves as a straightforward indicator, it may not capture all valid problem structures. As a result, we risk excluding valid rephrased versions without question marks. Furthermore, there are instances where the model generates empty strings, leading to the absence of rephrased versions for certain samples.  It's noteworthy that, as shown in Table \ref{tab:numbers}, occurrences of these conditions are relatively low, especially in the SVAMP dataset.

        \item Diversity Enhancement: Rephrased examples identical to the original ones are omitted to enhance diversity in the augmented dataset. 
        In step 1, we employ an LLM to generate diverse variations of the original questions. However, due to the inherent nature of the model, there were cases where the rephrased produced questions were identical to the original ones.
        Therefore, we implemented an automatic process to identify and remove these duplicate examples to address this issue. If an exact match was found, indicating redundancy in the augmented dataset, those examples were eliminated to ensure the diversity and uniqueness of the dataset.
    \end{itemize}
    The filtration process is done automatically. There is no involvement of humans. 

    \item \textbf{Step 3: Numerical Modification:}
    The subsequent phase of the algorithm focuses on numerical modification to introduce variability in the dataset. This involves random replacement of numerical values, where randomly generated integers replace integer values, and float numbers are replaced by randomly generated float numbers. Crucially, to maintain consistency, the algorithm ensures that modifications applied to numerical values are consistently propagated to the corresponding equations and answers. Equations and answers are reconstructed to preserve the logical and mathematical integrity of the examples.

    Note that, changing the numerical values is significant for several reasons. While introducing quantity-agnostic tags/tokens could be an alternative approach, changing the numerical values directly offers several advantages.
    Firstly, altering the numerical values allows for generating a more diverse set of problem equations/answers. This variation is crucial for training models to handle various scenarios.
    Secondly, changing the numerical values helps prevent the model from memorizing specific samples, equations/answers and encourages it to learn underlying patterns and relationships. This enables better generalization to unseen data and improves the model's robustness.
    While quantity-agnostic tags/tokens may offer some advantages, changing the numerical values directly remains a salient approach for enhancing the diversity, generalization, and coherence of generated MWPs.
\end{itemize}

The augmented dataset comprises the modified examples with the original examples. Through this new approach, our study offers a nuanced and practical approach to enhance the quality of rephrased examples in MWPs. The prompts used to generate the rephrased versions of the problems are provided in the appendices for reproducibility purposes (see Appendix \ref{secA1}).

\subsection{Rule-Based: Reversing Question}
Liu et al. introduced the concept of using inverted versions of problem statements in MWPs \cite{liu2021roda}.  Additionally, it is important to mention that this methodology has been investigated in other studies as well \cite{patel2021nlp, raiyan2023math}.
A similar approach is adopted in the previous study \cite{yigit2023exploring}. The actual problem text has been altered to make additional samples. The question and valid answer from the problem text generate a renewed supporting sentence. Then, a randomly picked sentence retaining a numerical value is converted into a new problem text, with the numerical value as the solution. This procedure forms new samples and assures contextual relevance. All possible equations related to the numerical values in the question text are created, increasing the data with various perspectives and helping the model to adapt to diverse challenges.

\subsection{Rule-Based: Question Replacement}
The previous study modified the question text to generate diverse problem types \cite{yigit2023exploring}. In this method, key phrases such as ``How many" and ``How much" in the original problem are replaced with ``What is x/y of," where x and y vary from 1 to 10. After this modification, a T5-Based model fine-tuned for grammar checking used to generate a modified version of the question text while maintaining consistency among the question, equation, and answer. This technique presented complexity by producing different texts for the same underlying question, diversifying the dataset, and preventing direct answer mapping. By incorporating such variations, the model becomes more robust at handling various problem formulations and answer types.

\subsection{Substitution: Synonym Replacement}
This method introduced synonyms from the NLTK WordNet to the original problems, adding semantic variations without varying the mathematical logic \cite{yigit2023exploring}. By randomly picking terms for replacement, the method increases the vocabulary of the problem text, contributing to a more diverse dataset. Using NLTK WordNet ensures that the selected synonyms are contextually pertinent, preserving the mathematical reasoning in the original question text. This strategy desires to create a dataset with various linguistic expressions while protecting the underlying mathematical structure in MWP solving.

\begin{table}[h]
\resizebox{\columnwidth}{!}{
\begin{tabular}{cccc}
\hline
 & Text & Equation & Answer \\ \hline
Original & \begin{tabular}[c]{@{}c@{}}Fred had 7 dimes in \colorbox{blue!30}{his bank} . His sister \colorbox{pink!30}{borrowed}\\ 3 of his  dimes . How many dimes does Fred have now?\end{tabular} & X=7-3 & 4 \\ \hline
\begin{tabular}[c]{@{}c@{}}Substitution: \\ Synonym \\ Replacement\end{tabular} & \begin{tabular}[c]{@{}c@{}}Fred   had 7 dimes in \colorbox{blue!30}{his depository financial institution}. \\ His  Sister \colorbox{pink!30}{lent} 3 of his dimes. How many dimes \\ does Fred have now?\end{tabular} & X=7-3 & 4 \\ \hline
\begin{tabular}[c]{@{}c@{}}Rule based: \\ Reversing \\ Question\end{tabular} & \begin{tabular}[c]{@{}c@{}}\colorbox{yellow!30}{Fred has 4 dimes now.} His sister borrowed 3 of \\ his dimes. \colorbox{red!30}{How many dimes in his bank did Fred have?}\end{tabular} & \colorbox{yellow!30}{4=X-3} & \colorbox{gray!30}{7} \\ \hline
\begin{tabular}[c]{@{}c@{}}Rule based: \\ Question\\ Replacement\end{tabular} & \begin{tabular}[c]{@{}c@{}}Fred  had 7 dimes in his bank. His sister borrowed \\ 3 of  his dimes. \colorbox{orange!30}{What is 9/10 of all the} dimes Fred \\ has now?\end{tabular} & X=(7-3)\colorbox{yellow!30}{*(9/10)} & \colorbox{gray!30}{3.6} \\ \hline
\begin{tabular}[c]{@{}c@{}}Rephrase with \\ In-Context \\ Learning\end{tabular} & \begin{tabular}[c]{@{}c@{}}\colorbox{cyan!30}{Fred initially had 23 dimes in his bank, but after} \\ \colorbox{cyan!30}{his sister borrowed 9 dimes,  how many dimes does Fred} \\ \colorbox{cyan!30}{have remaining?}\end{tabular} & \colorbox{yellow!30}{X=23-9} & \colorbox{gray!30}{14} \\ \hline
\end{tabular}
}
\caption{Examples from augmentation methods}\label{tab:augmentation_examples}
\end{table}

Table \ref{tab:augmentation_examples} shows examples of the proposed augmentation techniques on MWP. It includes the original text, equation, and answer, followed by the samples generated by proposed augmentation methods involving synonym replacement, rule based: reversing question, rule-based: question replacement, and in-context learning, and the corresponding equations and the answers. The modifications made to the original example are emphasized for each proposed augmentation method.

\section{Experiments} \label{sec:experiments}

\subsection{Datasets and Models}
\textbf{Datasets:}
The following datasets are used in the experiments:

\begin{itemize}
    \item \textbf{MAWPS-Single:}  MAWPS (MAth Word ProblemS) is a framework for building an online repository of MWPs, offering a diverse collection of MWPs along with their answers and equation templates \cite{koncel2016mawps}. MAWPS is designed to be a comprehensive and customizable resource for evaluating various algorithms.
    In the context of MAWPS and its sub-datasets, the term ``single equation" indicates that the problems involve mathematical expressions with a single unknown variable. The MAWPS-Single subset is designed to perform experiments or evaluations requiring or emphasizing one unknown problem type in MWPs. This subset comprises 1,987 MWPs, each featuring a single equation.

    \item \textbf{SVAMP:} SVAMP is a challenging dataset strategically collected from existing datasets with a specific focus on addressing the interesting observation that certain word problems can be solved without the complete problem text \cite{patel2021nlp}. 
    The creation of SVAMP involved the application of various modifications to seed examples sourced from the ASDiv-A dataset  \cite{miao2021diverse}. This choice was driven by the perceived higher quality and increased difficulty of ASDiv-A compared to the MAWPS dataset.
    This dataset comprised 3,138 problems in the training set and 1,000 problems in the test set. 
\end{itemize}

\ignore{
\begin{table}[]
\begin{tabular}{cccc}
\hline
Dataset & Question & Equation & Answer \\ \hline
MAWPS-Single & \begin{tabular}[c]{@{}c@{}}Mark\'s father gave him \$85. \\ Mark bought 10 books, each of \\ which cost \$5. How much money \\ does Mark have left?\end{tabular} & X=85-(10x5) & 35.0 \\ \hline
SVAMP & \begin{tabular}[c]{@{}c@{}}There were 8.0 friends playing\\  a video game online when 5.0 \\ players quit. If each player left \\ had 5.0 lives, how many lives did \\ they have total?\end{tabular} & X= (5.0x(8.0-5.0)) & 15.0 \\ \hline
\end{tabular}
\caption{Samples from MAWPS-Single and SVAMP datasets }\label{tab:examples}
\end{table}
}
These two datasets comprise the basis for the experiments, providing a varied and specialized set of MWPs for analyzing and evaluating algorithms in MWP solving.

\begin{table}[h]
\resizebox{\columnwidth}{!}{
\begin{tabular}{cccccccccc}
\hline
 & trainset & \begin{tabular}[c]{@{}c@{}}Question \\ Repl.\end{tabular} & \begin{tabular}[c]{@{}c@{}}\\ Reversing \\ Question\end{tabular} & \begin{tabular}[c]{@{}c@{}}Synonym\\ Repl.\end{tabular} & \begin{tabular}[c]{@{}c@{}}Rephrase with\\ In-Context \\ Learning\end{tabular} & \begin{tabular}[c]{@{}c@{}}Combined\\ V1\end{tabular} & \begin{tabular}[c]{@{}c@{}}Combined\\ V2\end{tabular} & \begin{tabular}[c]{@{}c@{}}Combined\\ V3\end{tabular} & \begin{tabular}[c]{@{}c@{}}Combined\\ V4\end{tabular} \\ \hline
\begin{tabular}[c]{@{}c@{}}asdiv-a\_\\ SVAMP\end{tabular} & 3138 & 5998 & 5553 & 6274 & 5744 & 11545 & 16810 & 14244 & 19509 \\ \hline
MAWPS-Single & 1589 & 3043 & 2557 & 3178 & 2996 & 5599 & 8022 & 7080 & 9503 \\ \hline
\end{tabular}
}
\caption{Number of training instances utilized for the proposed methods across the two datasets} \label{tab:numbers}
\end{table}

In addition to the previously mentioned augmentation sets, including the original examples, 4 additional trainsets have been constructed: Combined V1, Combined V2, Combined V3, and Combined V4. These sets are formulated by combining the samples generated via various augmentation methods. 
\begin{itemize}
    \item Combined V1 contains instances generated by Rule-Based procedures such as Question Replacement and Reversing Questions, along with the actual training samples. 
    \item Combined V2 comprises instances created by Rule-Based methods and actual training samples. Then, Substitution: Synonym Replacement approach is applied to all those examples, so it comprises more instances than Combined V1. 
    \item Combined V3 contains all examples from Combined V1 and adds examples generated by the in-context learning augmentation approach, ensuring no duplication of the original training data.
    \item Finally, Combined V4 includes all examples from Combined V3 and incorporates additional samples generated through the in-context augmentation approach, also ensuring that there is no duplication of the original training data.\end{itemize}

Table \ref{tab:numbers} shows the number of training instances utilized for the proposed methods across the two datasets, including the combined sets.
Having diversified sets aims to enhance the robustness and generalization of the training data for improved performance in MWP solving.

\begin{table}[]
\resizebox{\columnwidth}{!}{
\begin{tabular}{ccccccc}
\hline
Model & Year & Type & Description & Encoder & Decoder & \begin{tabular}[c]{@{}c@{}}Pretrained \\ model\end{tabular} \\ \hline
RNNVAE \cite{zhang2016variational} & 2016 & Seq2Seq & \begin{tabular}[c]{@{}c@{}}Hybrid   variational encoder-decoder \\ system integrating VAE and RNN principles.   \\ Leverages VAE strengths in text modeling \\ and sequential  processing abilities of RNNs.\end{tabular} & LSTM & LSTM & - \\ \hline
DNS \cite{wang2017deep} & 2017 & Seq2Seq & \begin{tabular}[c]{@{}c@{}}Deep   neural solver for MWPs. \\ Utilizes a hybrid model with a \\ similarity-based  retrieval model for \\ performance improvements.\end{tabular} & GRU & LSTM & - \\ \hline
MathEN \cite{wang2018translating} & 2018 & Seq2Seq & \begin{tabular}[c]{@{}c@{}}Addresses   the issue of multiple correct\\  equations in Seq2Seq models for MWPs.   \\ Introduces an equation normalization\\  method.\end{tabular} & BiLSTM & LSTM & - \\ \hline
GTS \cite{xie2019goal} & 2019 & Seq2Tree & \begin{tabular}[c]{@{}c@{}}Goal-driven   tree-structured model for MWP\\  solving. Uses two-layer gated feed forward\\    networks for goal decomposition.\end{tabular} & GRU & TreeDecoder & - \\ \hline
RobertaGen \cite{liu2019roberta} & 2019 & Pre-trained & \begin{tabular}[c]{@{}c@{}}Advanced   language model based\\  on RoBERTa. Specializes in MWPs excelling\\  in contextual   and semantic understanding.\end{tabular} & RoBERTa & Transformer & RoBERTa \\ \hline
Saligned \cite{chiang2018semantically} & 2019 & Seq2Seq & \begin{tabular}[c]{@{}c@{}}Neural   encoder-decoder framework \\ for MWP solving. Focuses on the semantic\\  meanings   of symbols in the text.\end{tabular} & BiLSTM & LSTM & - \\ \hline
Graph2Tree \cite{zhang2020graph} & 2020 & Graph2Tree & \begin{tabular}[c]{@{}c@{}}Deep   learning architecture for improved \\ solution equation generation in MWPs. \\ Uses Quantity Cell and Quantity Comparison \\ Graphs for representation.\end{tabular} & \begin{tabular}[c]{@{}c@{}}LSTM   \\ + \\ Graph \\ Convolutional\\  Networks\end{tabular} & TreeDecoder & - \\ \hline
SAUSolver \cite{qin2020semantically} & 2020 & Seq2Tree & \begin{tabular}[c]{@{}c@{}}Semantically aligned universal tree structured\\  solver. Encoder-decoder framework   generating \\ a universal expression tree based on semantic \\ meanings.\end{tabular} & GRU & TreeDecoder & - \\ \hline
MWPBert \cite{liang2021mwp} & 2021 & Seq2Tree & \begin{tabular}[c]{@{}c@{}}Addresses   the issue of number representation\\  in MWP solving. BERT-based model injecting\\    numerical properties into symbolic placeholders.\end{tabular} & Bert & TreeDecoder & Bert \\ \hline
\end{tabular}
}
\caption{Comparison of MAWP Solving Models}\label{tab:models}
\end{table}

\textbf{Models:}
This study assesses and compares diverse neural models for MWP solving. A comparative table of these models is given in Table \ref{tab:models}. Each model is designed to overcome distinct challenges in the domain. GTS presents a novel tree-structured model that works goal-driven, recursively decomposing the problem into sub-goals until getting a known quantity. 
DNS converts MWPs directly into equation templates using an RNN, improving performance with a hybrid model integrating a similarity-based retrieval model. RobertaGen, built upon RoBERTa, makes a powerful tool for a nuanced understanding of mathematical complexities. RNNVAE leverages directions from variational autoencoders and RNNs, incorporating VAE's text modeling with RNN's sequential processing to obtain accurate MWP answers.

Moreover, Graph2Tree addresses challenges in capturing relationships between quantities in tree-based neural models. MathEN focuses on the issue of multiple correct equations in MWPs. It introduces an ensemble model integrating the strengths of individual Seq2Seq models. Saligned utilizes a neural encoder-decoder framework focused on the semantic meanings of symbols in the text. MWPBert, a BERT-based model, addresses number representation challenges by injecting numerical properties into symbolic placeholders. SAUSolver generates universal expression trees based on the semantic meanings of previously generated symbols. 

The models employed in this study can be classified based on their architectures: GTS, MWPBert, and SAUSolver serve as Seq2Tree models; RNNVAE, DNS, MathEN, and Saligned utilize Seq2Seq architectures; Graph2Tree is specifically designed as a Graph2Tree model, and RobertaGen stands out as a pre-trained model. This categorization emphasizes the study's diverse architectural strategies for MWP solving.

\subsection{Experimental Results}
Our experiments mainly focus on the MAWPS-Single and SVAMP datasets, which are widely used one-unknown MWP datasets.
The performances of the conducted experiments on MAWPS-Single and SVAMP datasets are illustrated in Table \ref{tab:mawps_single} and \ref{tab:svamp}. Rows and columns describe models and presented augmentation strategies. Results are underlined and bold for instances where the augmentation technique performs less than the reproduced performances. Further, the highest two performances on the test set are highlighted in red.

We employ two evaluation metrics, equation accuracy and answer accuracy, to assess the performance of the models. Equation accuracy determines whether the predicted solution equation matches the expected equation precisely. Moreover, answer accuracy evaluates the exact matches between predicted and expected answers. By utilizing both metrics, we comprehensively assess the effectiveness of our models in generating accurate solutions to MWPs. In addition, our choice of accuracy for comparison aligns with the evaluation metric used by state-of-the-art baseline models in the field. This ensures a direct and fair comparison between our proposed methods and existing approaches.


\begin{table}[h]
\resizebox{\columnwidth}{!}{
\begin{tabular}{ccccccccccccc}
\hline
\multicolumn{3}{c}{MAWPS-Single} & \begin{tabular}[c]{@{}c@{}}Acc \\ in \cite{lan2022mwptoolkit}\end{tabular} & Reproduced & \begin{tabular}[c]{@{}c@{}}Question \\ Repl.\end{tabular} & \begin{tabular}[c]{@{}c@{}}Reversing \\ Question\end{tabular} & \begin{tabular}[c]{@{}c@{}}Synonym \\ Repl.\end{tabular} & \begin{tabular}[c]{@{}c@{}}Rephrase\\ in\\ context \\ learning\end{tabular} & \begin{tabular}[c]{@{}c@{}}Combined\\ V1\end{tabular} & \begin{tabular}[c]{@{}c@{}}Combined\\ V2\end{tabular} & \begin{tabular}[c]{@{}c@{}}Combined\\ V3\end{tabular} & \begin{tabular}[c]{@{}c@{}}Combined\\ V4\end{tabular} \\ \hline
 &  & \begin{tabular}[c]{@{}c@{}}equ\\ acc\end{tabular} & 78,9 & 79,4 & 79,4 & 79,4 & 79,4 & 79,9 & {\color[HTML]{FF0000} 81,4} & 80,4 & 79,9 & {\color[HTML]{FF0000} 80,9} \\ 
 & \multirow{-2}{*}{DNS} & \begin{tabular}[c]{@{}c@{}}val\\ acc\end{tabular} & 86,3 & 88,4 & \textbf{\underline{87,9}} & \textbf{\underline{87,9}} & \textbf{\underline{87,9}} & \textbf{\underline{87,9}} & {\color[HTML]{FF0000} 88,9} & {\color[HTML]{FF0000} 88,9} & \textbf{\underline{87,4}} & \textbf{\underline{87,4}} \\ 
 &  & \begin{tabular}[c]{@{}c@{}}equ\\ acc\end{tabular} & 85,9 & 84,4 & {\color[HTML]{FF0000} 87,9} & 85,9 & 86,9 & 85,9 & {\color[HTML]{FF0000} 87,9} & 86,9 & 86,4 & 86,4 \\ 
 & \multirow{-2}{*}{MathEN} & \begin{tabular}[c]{@{}c@{}}val\\ acc\end{tabular} & 86,4 & 85,9 & {\color[HTML]{FF0000} 88,9} & 86,9 & 87,9 & 86,9 & {\color[HTML]{FF0000} 88,9} & 87,9 & 87,4 & 87,4 \\ 
 &  & \begin{tabular}[c]{@{}c@{}}equ\\ acc\end{tabular} & 86 & 77,4 & 78,4 & 79,9 & 78,4 & 77,4 & {\color[HTML]{FF0000} 80,9} & 79,9 & {\color[HTML]{FF0000} 80,4} & 78,4 \\ 
 & \multirow{-2}{*}{Saligned} & \begin{tabular}[c]{@{}c@{}}val\\ acc\end{tabular} & 86,3 & 85,4 & 88,9 & 87,9 & 87,9 & 85,9 & {\color[HTML]{FF0000} 88,9} & 87,9 & {\color[HTML]{FF0000} 87,9} & 86,4 \\ 
 &  & \begin{tabular}[c]{@{}c@{}}equ\\ acc\end{tabular} & 79,8 & 87,9 & 88,9 & \textbf{\underline{87,4}} & \textbf{\underline{87,4}} & 87,9 & 88,9 & {\color[HTML]{FF0000} 89,4} & 87,9 & {\color[HTML]{FF0000} 89,4} \\ 
\multirow{-12}{*}{\rotatebox[origin=c]{90}{Seq2Seq}} & \multirow{-2}{*}{RNNVAE} & \begin{tabular}[c]{@{}c@{}}val\\ acc\end{tabular} & 88,2 & 88,9 & 89,9 & 88,9 & \textbf{88,4} & 88,9 & 89,9 & {\color[HTML]{FF0000} 90,5} & 88,9 & {\color[HTML]{FF0000} 89,9} \\ \hline
{\color[HTML]{1F2328} } &  & \begin{tabular}[c]{@{}c@{}}equ\\ acc\end{tabular} & - & 84,4 & 86,4 & \textbf{\underline{82,4}} & 86,4 & 86,4 & {\color[HTML]{FF0000} 86,9} & 85,9 & {\color[HTML]{FF0000} 86,9} & {\color[HTML]{FF0000} 88,4} \\ 
{\color[HTML]{1F2328} } & \multirow{-2}{*}{MWPBert} & \begin{tabular}[c]{@{}c@{}}val\\ acc\end{tabular} & - & 84,9 & 86,9 & \textbf{\underline{82,9}} & 86,9 & 86,9 & {\color[HTML]{FF0000} 87,4} & 86,4 & 86,9 & {\color[HTML]{FF0000} 88,4} \\ 
{\color[HTML]{1F2328} } &  & \begin{tabular}[c]{@{}c@{}}equ\\ acc\end{tabular} & 83,4 & 83,9 & 85,9 & {\color[HTML]{FF0000} 86,4} & 84,9 & 85,9 & 85,4 & 85,4 & 85,9 & {\color[HTML]{FF0000} 86,4} \\ 
{\color[HTML]{1F2328} } & \multirow{-2}{*}{SAUSolver} & \begin{tabular}[c]{@{}c@{}}val\\ acc\end{tabular} & 84 & 84,9 & 86,4 & {\color[HTML]{FF0000} 86,9} & 85,4 & 86,4 & 85,9 & 85,9 & 86,4 & {\color[HTML]{FF0000} 86,4} \\ 
{\color[HTML]{1F2328} } &  & \begin{tabular}[c]{@{}c@{}}equ\\ acc\end{tabular} & 83,5 & 83,9 & 86,9 & 84,9 & 85,9 & {\color[HTML]{FF0000} 87,4} & 85,4 & 84,9 & 85,4 & {\color[HTML]{FF0000} 87,4} \\ 
\multirow{-12}{*}{{\color[HTML]{1F2328} \rotatebox[origin=c]{90}{Seq2Tree}}} & \multirow{-2}{*}{GTS} & \begin{tabular}[c]{@{}c@{}}val\\ acc\end{tabular} & 84,1 & 84,4 & 87,4 & 85,4 & 86,4 & {\color[HTML]{FF0000} 87,9} & 85,9 & 85,4 & 85,9 & {\color[HTML]{FF0000} 87,9} \\ \hline
{\color[HTML]{1F2328} } &  & \begin{tabular}[c]{@{}c@{}}equ\\ acc\end{tabular} & 84,9 & 84,9 & 86,4 & 84,9 & 87,4 & 86,4 & 86,9 & 86,4 & {\color[HTML]{FF0000} 88,4} & {\color[HTML]{FF0000} 87,9} \\ 
\multirow{-4}{*}{{\color[HTML]{1F2328} Graph2Tree}} & \multirow{-2}{*}{Graph2Tree} & \begin{tabular}[c]{@{}c@{}}val\\ acc\end{tabular} & 85,6 & 85,4 & 86,9 & 85,4 & 87,9 & 85,4 & 87,4 & 86,9 & {\color[HTML]{FF0000} 89,4} & {\color[HTML]{FF0000} 88,9} \\ \hline
{\color[HTML]{1F2328} } &  & \begin{tabular}[c]{@{}c@{}}equ\\ acc\end{tabular} & 80,8 & 83,4 & {\color[HTML]{FF0000} 85,9} & 84,9 & 84,4 & 84,4 & 83,9 & 84,4 & {\color[HTML]{FF0000} 85,9} & 85,4 \\ 
\multirow{-2}{*}{{\color[HTML]{1F2328} Pre-trained}} & \multirow{-2}{*}{RobertaGen} & \begin{tabular}[c]{@{}c@{}}val\\ acc\end{tabular} & 88,4 & 84,9 & {\color[HTML]{FF0000} 86,9} & 85,9 & 84,9 & 85,4 & \textbf{\underline{84,4}} & 85,9 & {\color[HTML]{FF0000} 86,9} & 85,9 \\ \hline
\end{tabular}
}
\caption{Experiments on MAWPS-Single dataset}\label{tab:mawps_single}
\end{table}

\begin{table}[h]
\resizebox{\columnwidth}{!}{
\begin{tabular}{ccccccccccccc}
\hline
\multicolumn{3}{c}{SVAMP} & \begin{tabular}[c]{@{}c@{}}Acc \\ in \cite{lan2022mwptoolkit}\end{tabular} & Reproduced & \begin{tabular}[c]{@{}c@{}}Question \\ Repl.\end{tabular} & \begin{tabular}[c]{@{}c@{}}Reversing \\ Question\end{tabular} & \begin{tabular}[c]{@{}c@{}}Synonym \\ Repl.\end{tabular} & \begin{tabular}[c]{@{}c@{}}Rephrase\\ in\\ context \\ learning\end{tabular} & \begin{tabular}[c]{@{}c@{}}Combined\\ V1\end{tabular} & \begin{tabular}[c]{@{}c@{}}Combined\\ V2\end{tabular} & \begin{tabular}[c]{@{}c@{}}Combined\\ V3\end{tabular} & \begin{tabular}[c]{@{}c@{}}Combined\\ V4\end{tabular} \\ \hline
 &  & \begin{tabular}[c]{@{}c@{}}equ\\ acc\end{tabular} & 22.1 & 19 & {\color[HTML]{FF0000} 23,7} & 20,2 & 23,1 & 21,2 & 23,9 & {\color[HTML]{FF0000} 23,6} & 22,5 & 23,3 \\ 
 & \multirow{-2}{*}{DNS} & \begin{tabular}[c]{@{}c@{}}val\\ acc\end{tabular} & 24.2 & 22,6 & {\color[HTML]{FF0000} 26,8} & 23,6 & 26,8 & 23,5 & 26,9 & {\color[HTML]{FF0000} 26,5} & 25,2 & 26,2 \\ 
 &  & \begin{tabular}[c]{@{}c@{}}equ\\ acc\end{tabular} & 21.8 & 23,7 & 25,1 & 24,8 & 24,3 & \textbf{\underline{23,2}} & 25,1 & 25,7 & {\color[HTML]{FF0000} 26,5} & {\color[HTML]{FF0000} 26,3} \\ 
 & \multirow{-2}{*}{MathEN} & \begin{tabular}[c]{@{}c@{}}val\\ acc\end{tabular} & 25.0 & 24,3 & 25,6 & 25,2 & 24,9 & \textbf{\underline{23,9}} & 25,6 & 26,3 & {\color[HTML]{FF0000} 27,3} & {\color[HTML]{FF0000} 27} \\ 
 &  & \begin{tabular}[c]{@{}c@{}}equ\\ acc\end{tabular} & 23.9 & 24,1 & 24,3 & \textbf{\underline{21,6}} & 24,5 & 24,5 & 25 & {\color[HTML]{FF0000} 26} & \textbf{\underline{24}} & {\color[HTML]{FF0000} 25,5} \\ 
 & \multirow{-2}{*}{Saligned} & \begin{tabular}[c]{@{}c@{}}val\\ acc\end{tabular} & 26.1 & 27,2 & \textbf{\underline{27}} & \textbf{\underline{24,6}} & 27,6 & 27,6 & 28,5 & {\color[HTML]{FF0000} 28,5} & 27,9 & {\color[HTML]{FF0000} 28,7} \\ 
 &  & \begin{tabular}[c]{@{}c@{}}equ\\ acc\end{tabular} & 23.2 & 21,7 & 24,1 & 23,4 & 22,6 & 23,6 & {\color[HTML]{FF0000} 24,6} & 23,4 & 25,2 & {\color[HTML]{FF0000} 24,7} \\ 
\multirow{-12}{*}{\rotatebox[origin=c]{90}{Seq2Seq}} & \multirow{-2}{*}{RNNVAE} & \begin{tabular}[c]{@{}c@{}}val\\ acc\end{tabular} & 25.9 & 25,4 & 27,2 & 26,2 & 26,7 & 26,9 & {\color[HTML]{FF0000} 28,3} & 27,4 & 28,4 & {\color[HTML]{FF0000} 28,5} \\ \hline
{\color[HTML]{1F2328} } &  & \begin{tabular}[c]{@{}c@{}}equ\\ acc\end{tabular} & - & 24,4 & 25,1 & \textbf{\underline{24,2}} & 26,6 & {\color[HTML]{FF0000} 28,5} & 26 & 27,4 & 27,4 & {\color[HTML]{FF0000} 27,7} \\ 
{\color[HTML]{1F2328} } & \multirow{-2}{*}{MWPBert} & \begin{tabular}[c]{@{}c@{}}val\\ acc\end{tabular} & - & 26,6 & 28,5 & \textbf{\underline{26}} & 29,4 & {\color[HTML]{FF0000} 32,8} & 28,3 & 30,1 & 31,1 & {\color[HTML]{FF0000} 31,4} \\ 
{\color[HTML]{1F2328} } &  & \begin{tabular}[c]{@{}c@{}}equ\\ acc\end{tabular} & 27.1 & 25,1 & 26 & \textbf{\underline{24,3}} & \textbf{\underline{25}} & 25,7 & {\color[HTML]{FF0000} 26,9} & 25,8 & 26,3 & {\color[HTML]{FF0000} 27} \\ 
{\color[HTML]{1F2328} } & \multirow{-2}{*}{SAUSolver} & \begin{tabular}[c]{@{}c@{}}val\\ acc\end{tabular} & 29.7 & 27,6 & 28,3 & \textbf{\underline{27}} & 28,3 & 28,5 & {\color[HTML]{FF0000} 30,2} & 28,4 & {\color[HTML]{FF0000} 29,1} & 29 \\ 
{\color[HTML]{1F2328} } &  & \begin{tabular}[c]{@{}c@{}}equ\\ acc\end{tabular} & 25.6 & 25,3 & {\color[HTML]{FF0000} 26,7} & \textbf{\underline{24,2}} & 25,5 & 25,9 & 26 & 25,5 & 26,3 & {\color[HTML]{FF0000} 27} \\ 
\multirow{-12}{*}{{\color[HTML]{1F2328} \rotatebox[origin=c]{90}{Seq2Tree}}} & \multirow{-2}{*}{GTS} & \begin{tabular}[c]{@{}c@{}}val\\ acc\end{tabular} & 29.1 & 28,5 & {\color[HTML]{FF0000} 29} & \textbf{27,2} & 28,6 & 28,9 & 29,1 & 28,4 & 29,4 & {\color[HTML]{FF0000} 29} \\ \hline
{\color[HTML]{1F2328} } &  & \begin{tabular}[c]{@{}c@{}}equ\\ acc\end{tabular} & 31.6 & 32,5 & 33,6 & \textbf{\underline{31,4}} & 32,6 & {\color[HTML]{FF0000} 34} & {\color[HTML]{FF0000} 34} & 33,2 & \textbf{\underline{31}} & \textbf{\underline{31,4}} \\ 
\multirow{-2}{*}{{\color[HTML]{1F2328} Graph2Tree}} & \multirow{-2}{*}{Graph2Tree} & \begin{tabular}[c]{@{}c@{}}val\\ acc\end{tabular} & 35.0 & 35,3 & 36,7 & \textbf{\underline{34,7}} & 35,7 & {\color[HTML]{FF0000} 35} & {\color[HTML]{FF0000} 36,8} & 36,6 & \textbf{\underline{33,2}} & \textbf{\underline{33}} \\ \hline
{\color[HTML]{1F2328} } &  & \begin{tabular}[c]{@{}c@{}}equ\\ acc\end{tabular} & 27.9 & 20,9 & 21,9 & \textbf{\underline{20,8}} & {\color[HTML]{FF0000} 22,1} & 21,8 & 21,9 & 21,5 & {\color[HTML]{FF0000} 23,4} & \textbf{\underline{20,6}} \\ 
\multirow{-2}{*}{{\color[HTML]{1F2328} Pre-trained}} & \multirow{-2}{*}{RobertaGen} & \begin{tabular}[c]{@{}c@{}}val\\ acc\end{tabular} & 30.3 & 24 & 24,4 & \textbf{\underline{23,3}} & {\color[HTML]{FF0000} 24,9} & 25,1 & 25 & 24 & {\color[HTML]{FF0000} 26,3} & 24,4 \\ \hline
\end{tabular}
}
\caption{Experiments on SVAMP dataset}\label{tab:svamp}
\end{table}

Our study employed proposed augmentation methods across various models, including 4 Seq2Seq, 3 Seq2Tree, 1 Graph2Tree, and 1 RobertaGen model. 
In our analysis, we compared the obtained performances with the baseline in \cite{lan2022mwptoolkit}. Through a comparison of these performances, the following outcomes can be inferred.

\begin{table}[]
\resizebox{\columnwidth}{!}{
\begin{tabular}{cccccccccc}
\hline
 &  & \begin{tabular}[c]{@{}c@{}}Question\\ Repl.\end{tabular} & \begin{tabular}[c]{@{}c@{}}Reversing\\ Question\end{tabular} & \begin{tabular}[c]{@{}c@{}}Synonym\\ Repl.\end{tabular} & \begin{tabular}[c]{@{}c@{}}Rephrase \\ with\\ In-Context \\ Learning\end{tabular} & \begin{tabular}[c]{@{}c@{}}Combined\\      V1\end{tabular} & \begin{tabular}[c]{@{}c@{}}Combined\\      V2\end{tabular} & \begin{tabular}[c]{@{}c@{}}Combined\\      V3\end{tabular} & \begin{tabular}[c]{@{}c@{}}Combined\\      V4\end{tabular} \\ \hline
 & \begin{tabular}[c]{@{}c@{}}equ\\ acc\end{tabular} & 9/9 & {\color[HTML]{FF0000} 3/9} & 8/9 & 8/9 & 9/9 & 9/9 & 7/9 & 7/9 \\ 
\multirow{-2}{*}{SVAMP} & \begin{tabular}[c]{@{}c@{}}val\\ acc\end{tabular} & 8/9 & {\color[HTML]{FF0000} 3/9} & 9/9 & 8/9 & 9/9 & 9/9 & 8/9 & 7/9 \\ \hline
 & \begin{tabular}[c]{@{}c@{}}equ\\ acc\end{tabular} & 9/9 & 7/9 & 7/9 & 9/9 & 9/9 & 9/9 & 9/9 & 9/9 \\ 
\multirow{-2}{*}{MAWPS-Single} & \begin{tabular}[c]{@{}c@{}}val\\ acc\end{tabular} & 8/9 & 7/9 & 8/9 & 8/9 & 8/9 & 9/9 & 8/9 & 8/9 \\ \hline
\end{tabular}
}
\caption{Performance Comparison of the Models with Various Augmentation Techniques}\label{tab:compare_performances}
\end{table}

\begin{itemize}
    \item In general, the proposed data augmentation methods performed better across nine distinct models.
    \item The combined augmentation sets mostly outperformed other individual methods.
   \item On the MAWPS-Single dataset, 
    \begin{itemize}
        \item The Seq2Seq models revealed their highest performances mainly over Combined V1.
        \item Seq2Tree, Graph2Tree, and pre-trained models performed better when trained on Combined V3. 
    \end{itemize}
    \item Similarly, for the Svamp dataset, the superiority of combined datasets, specifically Combined v4, was evident as it consistently achieved the highest performances.
\end{itemize}

Moreover, Table \ref{tab:compare_performances} provides a detailed comparison of the performance of different augmentation approaches across datasets and evaluation metrics. Each row corresponds to a specific dataset with an evaluation metric, while the columns represent different augmentation approaches employed.
Each cell in the table contains a value indicating how each augmentation approach performs relative to baseline methods. For instance, the value ``3/9" in a cell indicates that the corresponding augmentation approach outperformed the baseline in three out of nine models considered for evaluation.
According to this table:
    \begin{itemize}
        \item The models generally performed well with Rule Based: Question Replacement, Synonym Replacement, and augmentation with in-context learning approaches, achieving 7/9, 8/9, and 9/9 in both equation and value accuracy.
        \item Training with Combined V2, the models consistently demonstrated strong performance with 9/9 in both equation and value accuracy.
        \item Rule-based: Reversing Question method had a mixed impact on model performance. Of the 9 models examined, 3 displayed superior performance, outperforming the reproduced performances. This observation underscores the importance of understanding the impact of augmentation techniques, on model outcomes for the SVAMP dataset.
    \end{itemize}

In summary, experimental results highlight the significance of dataset composition and augmentation strategies on various model performances.

\section{Discussion} \label{sec:discussion}

Our proposed methods address several challenges inherent in MWPs augmentation.
Synonym replacement ensures that the meaning and context of the original problem statements are maintained by introducing variation while preserving the semantic meaning of the original problems. Similarly, rule-based question replacement and reversing methodologies are applied strategically to ensure the generation of diverse and contextually relevant problem statements, thus maintaining the semantic integrity of the dataset. The in-context learning based approach ensures that generated problem statements are contextually relevant, preserving semantic integrity.
Moreover, our proposed methods introduce diversity into the dataset, thereby reducing the risk of model bias. By incorporating diverse synonyms and applying predefined rules, we expose models to a broader range of examples, reducing the risk of bias in the training data, especially on concatenated sets. Similarly, by leveraging the capabilities of the Llama-7b language model in in-context learning based approach, the risk of model bias is reduced.

The methodologies of Synonym Replacement, Rule-Based Question Replacement, and Rule-Based Reversing Question are employed in a controlled manner to maintain uniformity in the textual information while preserving mathematical relationships. These methods ensure that the semantic integrity of the dataset is preserved by introducing variation while maintaining coherence in the problem statements. Additionally, these techniques apply no numerical modifications, ensuring that the numerical information remains consistent across the dataset.
In in-context learning approach, numerical modifications are applied, but so in a controlled manner. This ensures that changes to numerical information are uniform and coherent, preserving the underlying mathematical relationships.

In in-context learning based approach, we utilize prompt-based generation using an LLM. These models are trained on extensive datasets, which helps ensure their semantic integrity and robustness. While we acknowledge the importance of consistency in the rephrased statements with the equations and answers, we rely on the inherent capabilities of the LLM to maintain semantic coherence during the generation process. Additionally, the problem texts provided to the models are simple, enhancing the likelihood of consistent and accurate rephrased statements aligning with the equations and answers.

To ensure that our proposed augmentation approaches are not solely relying on shallow heuristics or memorizing augmented sample templates to achieve improved accuracy, 
\begin{itemize}
    \item The proposed methods are evaluated on multiple datasets to assess their ability to generalize across different problem domains and variations.
    \item The experiments are conducted on 9 baseline models, which serve as benchmarks for comparison. This allows us to assess the relative performance of our approaches and verify that any improvements in accuracy are not simply due to overfitting or memorization.
    \item The diversity of our training data is enhanced by concatenating examples generated by different proposed methods. Increasing the variety of training examples encourages the models to learn robust and generalizable patterns, rather than relying on memorized templates or shallow heuristics.

\end{itemize}

We have also thoroughly assessed the efficacy of our proposed in-context learning approach using the MWPBert model. The examples provided in Table \ref{tab:mwp-samples} demonstrate a set of MWPs samples that were initially incorrectly solved by the MWPBert model prior to its training with augmented samples. However, after training with the augmented samples generated through the in-context learning approach, the model successfully generates correct answers. This improvement provides the claim that our augmentation approach enhances the model's reasoning ability when solving MWPs.

\begin{table}[h]
\caption{The MWP samples were incorrectly solved by MWPBert, correctly solved after training with in-context learning based approach}\label{tab:mwp-samples}
\centering
\begin{tabular}{|l|l|l|}
\hline
Question &
  Equation &
  Answer \\ \hline
\begin{tabular}[c]{@{}l@{}}There are 20 different books in the 'crazy silly\\ school' series. If you are yet to read 5 of the \\ books, how many books have you already read?\end{tabular} &
  X=(20.0 - 5.0) &
  15 \\ \hline
\begin{tabular}[c]{@{}l@{}}Paul got a box of 440 crayons for his birthday. \\ During the school year he gave 111 crayons to\\ his friends while he lost 106 crayons. How many\\ crayons did he have left?\end{tabular} &
  X=(440.0 - (111.0 + 106.0)) &
  223 \\ \hline
\begin{tabular}[c]{@{}l@{}}Bobby had 21 pieces of candy. He ate 5 pieces\\ of candy.  Then he ate 9 more. How many pieces\\ of candy does he still have left?\end{tabular} &
  X=(21.0 - (5.0 + 9.0)) &
  7 \\ \hline
\begin{tabular}[c]{@{}l@{}}Dan has \$4. He bought a candy bar for \$7\\ and a chocolate for \$6. How much money did\\ he spend buying the candy bar and chocolate?\end{tabular} &
  X=(7.0 + 6.0) &
  13 \\ \hline
\end{tabular}
\end{table}

\section{Conclusion and Future Work}\label{sec:conclusion}
In conclusion, addressing the challenges posed by MWP solving in NLP is critical for improving the performance of existing models. The need for improved generalization highlights the significance of this study's objective to enhance MWP datasets with high-quality data. 
While this study's primary focus is on English, the study points to the potential generalization of the presented ideas to other languages in MWP solving. Several data augmentation techniques have been introduced and tested on MAWPS-Single and SVAMP datasets. The experiment results demonstrated the progress in performance on both datasets. 

In the future, we aim to investigate adversarial MWPs as part of our ongoing efforts, to improve the robustness and performance of our approach.
We will also explore the effectiveness of our in-context learning based augmentation approach with various other LLMs such as GPT-3 (ada, cabbage, curie, davinci), GPT-3.5 etc., as part of our future work. 

\backmatter

\bmhead{Acknowledgments}

This research is supported by The Scientific and Technological Research Council of Turkey (TÜBİTAK) in part of the project with 120E100 grant number. G.Yigit is supported by TUBİTAK - BİDEB 2211/A national fellowship program for Ph.D. studies.

\bmhead*{Declarations}
The authors declare that they have no conflict of interest.

\bmhead{Data Availability}
Data sharing not applicable to this article as no datasets were generated or analyzed during the current study.

\begin{appendices}

\section{}\label{secA1}

\begin{table}[h]
\caption{Prompts employed to generate rephrased versions of problem texts within the MAWPs-Single dataset}
\label{tab:prompts_mawps}
\begin{tabular}{|l|}
\hline
\begin{tabular}[c]{@{}l@{}}Your task   is to rephrase the given texts while preserving the numerical values and\\  relationships inherent in the original statements.\end{tabular} \\ 
\begin{tabular}[c]{@{}l@{}}\textbf{Text:} A store had 27   coloring books in stock. They ended up putting them on sale \\ and getting rid   of 6 of them. The put the ones they still had onto shelves with 7 on \\ each   shelf. How many shelves did they use?\end{tabular} \\ 
\begin{tabular}[c]{@{}l@{}}\textbf{Rephrased:} In stock, there   were 27 coloring books at a local store. During a sale, \\ 6 were sold, and the   remaining ones were neatly arranged on shelves, with 7 books \\ on each shelf.   How many shelves were utilized for this arrangement?\end{tabular} \\ 
\begin{tabular}[c]{@{}l@{}}\textbf{Text:} Shawn has 13   blocks. Mildred has with 2 blocks. Mildred finds another 84. \\ How many blocks   does Mildred end with?\end{tabular} \\ 
\begin{tabular}[c]{@{}l@{}}\textbf{Rephrased:} Shawn possesses 13   blocks, while Mildred starts with 2. If Mildred \\ discovers an additional 84   blocks, how many blocks does Mildred have in total?\end{tabular} \\ 
\begin{tabular}[c]{@{}l@{}}\textbf{Text:} Melanie grew 139   turnips. Benny grew 113 turnips. How many turnips did \\ they grow in all ?\end{tabular} \\ 
\begin{tabular}[c]{@{}l@{}}\textbf{Rephrased:} Melanie cultivated   139 turnips, and Benny grew 113. How many turnips\\  did they grow in total?\end{tabular} \\ 
\begin{tabular}[c]{@{}l@{}}\textbf{Text:} A teacher had 29   worksheets to grade. If she graded 25, but then another 29 \\ were turned in,   how many worksheets would she have to grade?\end{tabular} \\ 
\begin{tabular}[c]{@{}l@{}}\textbf{Rephrased:} A teacher had 29   worksheets to grade. After grading 25, an additional \\ 29 were turned in. How   many more worksheets does the teacher need to grade?\end{tabular} \\ 
\begin{tabular}[c]{@{}l@{}}\textbf{Text:} A pet store has 6   bird cages. If each cage has 2 parrots and 7 parakeets in it, \\ how many birds   does the pet store have total?\end{tabular} \\ 
\begin{tabular}[c]{@{}l@{}}\textbf{Rephrased:} In a pet store,   there are 6 bird cages. If each cage contains 2 parrots and\\  7 parakeets, how   many birds are there in total?\end{tabular} \\ 
\begin{tabular}[c]{@{}l@{}}\textbf{Text:} A restaurant   served 6 cakes during lunch and 9 during dinner today . How \\ many cakes were   served today ?\end{tabular} \\ 
\begin{tabular}[c]{@{}l@{}}\textbf{Rephrased:} Today, a restaurant   served 6 cakes during lunch and 9 during dinner. \\ How many cakes were served   in total?\end{tabular} \\ 
\begin{tabular}[c]{@{}l@{}}\textbf{Text:} Jessica spent $   10.22 on a cat toy , and a cage cost her $ 11.73 . What was \\ the total cost of   Jessica 's purchases ?\end{tabular} \\ 
\begin{tabular}[c]{@{}l@{}}\textbf{Rephrased:} Jessica spent   $10.22 on a cat toy, and a cage cost her $11.73. What was \\ the total cost of   Jessica's purchases?\end{tabular} \\ 
\begin{tabular}[c]{@{}l@{}}\textbf{Text:} Sam had to wash 40   short sleeve shirts and 23 long sleeve shirts before school. \\ If he had only   washed 29 of them by the time school started, how many did he not wash?\end{tabular} \\ 
\begin{tabular}[c]{@{}l@{}}\textbf{Rephrased:} Sam had 40 short   sleeve shirts and 23 long sleeve shirts to wash before school.\\  If he had only   washed 29 of them by the time school started, how many shirts did he not\\    wash?\end{tabular} \\ 
\begin{tabular}[c]{@{}l@{}}\textbf{Text:} George had 30   dollars. For his birthday he got 16 more dollars but spent 38 on a \\ new game.   How much money does he have now?\end{tabular} \\ 
\begin{tabular}[c]{@{}l@{}}\textbf{Rephrased:} For his birthday,   George received $16 more but spent $38 on a new game. \\ How much money does he   have now?\end{tabular} \\ 
\begin{tabular}[c]{@{}l@{}}\textbf{Text:} A pet store had 41   siamese cats and 28 house cats. During a sale they sold 15 \\ cats. How many   cats do they have left?\end{tabular} \\ 
\begin{tabular}[c]{@{}l@{}}\textbf{Rephrased:}  In a pet   store, there were 41 Siamese cats and 28 house cats. During a sale, \\ 15 cats   were sold. How many cats are left in the store?\end{tabular} \\ 
\begin{tabular}[c]{@{}l@{}}\textbf{Text:} Greg and Sharon   own neighboring cornfields. Greg harvested 0.4 of an acre of\\  corn on Monday   and Sharon harvested 0.1 of an acre. How many more acres did Greg\\  harvest   than Sharon?\end{tabular} \\ 
\begin{tabular}[c]{@{}l@{}}\textbf{Rephrased:} Greg and Sharon,   owners of neighboring cornfields, embarked on their \\ harvest. Greg gathered   0.4 acres of corn on Monday, while Sharon reaped 0.1 acres. \\ What is the   difference in the number of acres Greg harvested compared to Sharon?\end{tabular} \\ 
\begin{tabular}[c]{@{}l@{}}\textbf{Text:} A painter needed   to paint 6 rooms in a building. Each room takes 5 hours to \\ paint. If he   already painted 2 rooms, how much longer will he take to paint the rest?\end{tabular} \\ 
\begin{tabular}[c]{@{}l@{}}\textbf{Rephrased:} Tasked with   painting a building's 6 rooms, a diligent painter spends 5 \\ hours on each.   Having completed 2 rooms, how much additional time will it take for\\  the   painter to finish the remaining ones?\end{tabular} \\ 
\begin{tabular}[c]{@{}l@{}}\textbf{Text:} The Spurs   basketball team has 22 players. Each player has 11 basketballs. How\\  many   basketballs do they have in all?\end{tabular} \\ 
\begin{tabular}[c]{@{}l@{}}\textbf{Rephrased:} The Spurs   basketball team boasts 22 players, each equipped with 11 \\ basketballs. What is   the total number of basketballs in their possession?\end{tabular} \\ 
\begin{tabular}[c]{@{}l@{}}\textbf{Text:} Haley has 63   magazines in her cabinet. She plans to send it to the recycling \\ office in   their area. If she places it in boxes which can contain 9 magazines, how \\ many   boxes will Hayley use?\end{tabular} \\ 
\begin{tabular}[c]{@{}l@{}}\textbf{Rephrased:} Haley intends to   send 63 magazines from her cabinet to the recycling \\ office. If she organizes   them into boxes, each capable of holding 9 magazines, how \\ many boxes will she   need?\end{tabular} \\ 
\begin{tabular}[c]{@{}l@{}}\textbf{Text:} Frank worked 8   hours on the first 4 days of the week. How many hours did \\ he work in all?\end{tabular} \\ 
\begin{tabular}[c]{@{}l@{}}\textbf{Rephrased:}  Throughout   the initial 4 days of the week, Frank devoted 8 hours each\\  day to work. What   is the total number of hours he worked during this period?\end{tabular} \\ 
\begin{tabular}[c]{@{}l@{}}\textbf{Text:} A restaurant served 6   cakes during lunch and 9 during dinner today . How many\\  cakes were served   today ?\end{tabular} \\ 
\textbf{Rephrased:} \\ \hline
\end{tabular}
\end{table}

\begin{table}[h]
\caption{Prompts employed to generate rephrased versions of problem texts within the SVAMP dataset}
\label{tab:prompts_SVAMP}
\begin{tabular}{|l|}
\hline
\begin{tabular}[c]{@{}l@{}}Your task is to rephrase the given texts while preserving the numerical values and\\ relationships inherent in the original statements.\end{tabular} \\ 
\begin{tabular}[c]{@{}l@{}}\textbf{Text:} While playing a   trivia game , Mike answered 3.0 questions correct in the first \\ half and 5.0   questions correct in the second half . If each question was worth 3.0 \\ points,what was his final score ?\end{tabular} \\ 
\begin{tabular}[c]{@{}l@{}}\textbf{Rephrased:} Engaged in a trivia   game, Mike accurately responded to 3.0 questions\\  in the initial half and 5.0   questions in the latter half. If each question carried a value\\  of 3.0 points, what total score did Mike achieve?\end{tabular} \\ 
\begin{tabular}[c]{@{}l@{}}\textbf{Text:} Sam invited 9.0   friends to a birthday party , but 6.0 could n't come . If he \\ wanted to buy   enough cupcakes so each person could have exactly 2.0 , how many\\  should he   buy ?\end{tabular} \\ 
\begin{tabular}[c]{@{}l@{}}\textbf{Rephrased:} Inviting 9.0   friends to a birthday celebration, Sam faced the absence\\  of 6.0 attendees. To   ensure each person could enjoy exactly 2.0 cupcakes, how \\ many should Sam   purchase?\end{tabular} \\ 
\begin{tabular}[c]{@{}l@{}}\textbf{Text:} Keith grew 29.0   cantelopes , Fred grew 16.0 cantelopes , and Jason grew\\  20.0 cantelopes. How   many cantelopes did they grow in total ?\end{tabular} \\ 
\begin{tabular}[c]{@{}l@{}}\textbf{Rephrased:} Keith, Fred, and   Jason cultivated 29.0, 16.0, and 20.0 cantaloupes, \\ respectively. What is the   combined count of cantaloupes grown by the three?\end{tabular} \\ 
\begin{tabular}[c]{@{}l@{}}\textbf{Text:} Ezra drew a white   line that was 7 inches long . Then he drew \\ a blue line that   was 3 inches long . How much longer was the white\\  line than   the blue line ?\end{tabular} \\ 
\begin{tabular}[c]{@{}l@{}}\textbf{Rephrased:} Creating drawings,   Ezra crafted a white line measuring 7 inches and\\  a blue line   of 3 inches. What is the difference in length between the   white and\\  blue lines?\end{tabular} \\ 
\begin{tabular}[c]{@{}l@{}}\textbf{Text:} I have a pet   golden retriever . Each year he gains 11.0 pounds . He is 8.0 \\ years old . How   many pounds does he weigh ?\end{tabular} \\ 
\begin{tabular}[c]{@{}l@{}}\textbf{Rephrased:} Caring for a golden   retriever, he accumulates 11.0 pounds annually. \\ At the age of 8.0 years, what   is the total weight of the golden retriever?\end{tabular} \\ 
\begin{tabular}[c]{@{}l@{}}\textbf{Text:} finally they had   to roam around 169 factories to make sure they are \\ throwing their wastes properly . if their group went to 69 factories and the second\\  went to 52 how   many factories remain unchecked ?\end{tabular} \\ 
\begin{tabular}[c]{@{}l@{}}\textbf{Rephrased:} To ensure proper   waste disposal, the team needed to inspect 169 \\ factories. If one group   covered 69 factories and another visited 52, how many\\  factories are yet to be   checked?\end{tabular} \\ 
\begin{tabular}[c]{@{}l@{}}\textbf{Text:} There are 96.0   oranges in a box . Jonathan takes 45.0 oranges . How many\\  are left ?\end{tabular} \\ 
\begin{tabular}[c]{@{}l@{}}\textbf{Rephrased:} Within a box, there   exist 96.0 oranges. Jonathan claims 45.0 of them. \\ What is the remaining   count?\end{tabular} \\ 
\begin{tabular}[c]{@{}l@{}}\textbf{Text:} james has 1222   balloons . amy has 513 balloons . how many more balloons\\  does james have than   amy ?\end{tabular} \\ 
\begin{tabular}[c]{@{}l@{}}\textbf{Rephrased:} Possessing 1222   balloons, James exceeds Amy's collection by how\\  many balloons, given that Amy   has 513?\end{tabular} \\ 
\begin{tabular}[c]{@{}l@{}}\textbf{Text:} A dealer pays   6000.0 dollars for a car . The dealer wants to make a profit\\  that is 25.0 \%   of the selling price . For how much should the dealer sell the car ?\end{tabular} \\
\begin{tabular}[c]{@{}l@{}}\textbf{Rephrased:} If a dealer invests   6000.0 dollars in acquiring a car and aims for a \\ profit constituting 25.0\% of   the selling price, what should be the selling price?\end{tabular} \\ 
\begin{tabular}[c]{@{}l@{}}\textbf{Text:} In fourth grade   there were 11.0 students at the start of the year . During the\\  year 6.0   students left and 42.0 new students came to school . How many students were\\    in fourth grade at the end ?\end{tabular} \\ 
\begin{tabular}[c]{@{}l@{}}\textbf{Rephrased:} Commencing fourth   grade with 11.0 students, the class experienced\\  the departure of 6.0 students   and the arrival of 42.0 new students. What is the final\\  count of students in   fourth grade?\end{tabular} \\ 
\textbf{Text:} If Joan bicycled   25.0 miles at 5.0 miles per hour , how long was Joan travelling ? \\ 
\begin{tabular}[c]{@{}l@{}}\textbf{Rephrased:}  Covering a   distance of 25.0 miles on a bicycle moving at 5.0 miles per \\ hour, what was   the duration of Joan's travel?\end{tabular} \\ 
\begin{tabular}[c]{@{}l@{}}\textbf{Text:} A furniture store   has a chair , originally priced at 78.0 dollars , on sale for 46.0\\  dollars .   What is the percent of decrease , rounded to the nearest tenth ?\end{tabular} \\ 
\begin{tabular}[c]{@{}l@{}}\textbf{Rephrased:} The furniture store   offers a chair initially valued at 78.0 dollars for a sale \\ price of 46.0   dollars. What is the percentage decrease, rounded to the nearest tenth?\end{tabular} \\ 
\begin{tabular}[c]{@{}l@{}}\textbf{Text:} mrs. hilt ran 3   miles on monday 2 miles on wednesday and 7 miles on friday.\\  how many total   miles did she run that week ?\end{tabular} \\ 
\begin{tabular}[c]{@{}l@{}}\textbf{Rephrased:} Covering 3 miles on   Monday, 2 miles on Wednesday, and 7 miles on\\  Friday, what was the overall   distance Mrs. Hilt ran during the week?\end{tabular} \\ 
\begin{tabular}[c]{@{}l@{}}\textbf{Text:} A petri dish   originally contained 600.0 bacteria . A scientist let the bacteria\\  grow and   now there are 8917.0 of them . How many more bacteria are there now ?\end{tabular} \\ 
\begin{tabular}[c]{@{}l@{}}\textbf{Rephrased:} Starting with 600.0   bacteria, the population in a petri dish increased \\ to 8917.0 due to growth.   What is the additional count of bacteria?\end{tabular} \\ 
\begin{tabular}[c]{@{}l@{}}\textbf{Text:} Ryan has 72.0   marbles and 17.0 blocks . If he shares the marbles \\ among 9.0 friends , how   many marbles does each friend get ?\end{tabular} \\ 
\begin{tabular}[c]{@{}l@{}}\textbf{Rephrased:} With a possession   of 72.0 marbles and 17.0 blocks, if Ryan \\ distributes the marbles equally   among 9.0 friends, what is the share per friend?\end{tabular} \\ 
\begin{tabular}[c]{@{}l@{}}\textbf{Text:} A restaurant served 6   cakes during lunch and 9 during dinner today . \\ How many cakes were served   today ?\end{tabular} \\ 
\textbf{Rephrased:} \\ \hline
\end{tabular}
\end{table}

\end{appendices}



\end{document}